\documentclass{llncs}

\usepackage[a4paper,includeheadfoot,top=1.65in,bottom=1.65in,left=1.73in,hcentering]{geometry}
\usepackage[pdftex]{graphicx}
\usepackage[sectionbib]{natbib}

\usepackage[utf8]{inputenc}
\usepackage{t1enc}
\usepackage{tikz-dependency}
\usepackage{qtree}
\usepackage{float}
\restylefloat{figure}
\usepackage{siunitx}
\usepackage{tikz-qtree}
\usepackage[english]{babel}
\usepackage{todonotes}
\usepackage{array}
\usepackage{booktabs}
\usepackage{multirow}
\usepackage{amssymb}
\usepackage{hyperref}
\hypersetup{plainpages=false,
            pageanchor=true,
            colorlinks=true,
            linkcolor=blue,
            citecolor=blue,
            anchorcolor=blue,
            menucolor=blue,
            filecolor=blue,
            urlcolor=blue,
            linktocpage=true,
            breaklinks=true,
            unicode=true}
\usepackage{soul}

\newcommand{\embert}{\texttt{emBERT}}
\newcommand{\emtsv}{\texttt{emtsv}}

\newcommand{\magyarlanc}{\texttt{magyarlanc}}
\newcommand{\spacy}{spaCy}
\newcommand{\udpipe}{UDPipe}
\newcommand{\stanza}{Stanza}
\newcommand{\huspacy}{HuSpaCy}
\newcommand{\huntoken}{HunToken}
\newcommand{\lemmy}{Lemmy}

\begin{document}

\pagestyle{myheadings}
\def\leftmark{{\rm XVIII. Magyar Sz\'am\'\i t\'og\'epes Nyelv\'eszeti Konferencia}}
\def\rightmark{{\rm Szeged, 2022. január 27-28.}}

\title{\huspacy: an industrial-strength Hungarian natural language processing toolkit}

\author{
György Orosz,
Zsolt Szántó, \break
Péter Berkecz,
Gergő Szabó,
Richárd Farkas\\
\institute{
    Institute of Informatics, University of Szeged\break
    2. Árpád tér, Szeged, Hungary\\
}
\email{gyorgy@orosz.link}\break
\email{\{szantozs,rfarkas\}@inf.u-szeged.hu}\break
}

\maketitle

\begin{abstract}
Although there are a couple of open-source language processing pipelines available for Hungarian, none of them satisfies the requirements of today’s NLP applications. 
A language processing pipeline should consist of close to state-of-the-art lemmatization, morphosyntactic analysis, entity recognition and word embeddings. 
Industrial text processing applications have to satisfy non-functional software quality requirements, what is more, frameworks supporting multiple languages are more and more favored.
This paper introduces \huspacy, an industry-ready Hungarian language processing toolkit. 
The presented tool provides components for the most important basic linguistic analysis tasks. It is open-source and is available under a permissive license. 
Our system is built upon spaCy’s NLP components resulting in an easily usable, fast yet accurate application.
Experiments confirm that HuSpaCy has high accuracy while maintaining resource-efficient prediction capabilities. 
\end{abstract}

\section{Introduction}

Basic natural language processing tasks such as tokenization, sentence splitting, part-of-speech tagging, lemmatization, dependency parsing and named entity recognition are amongst the most widely studied problems in natural language processing. Several text analysis applications have been developed during the last decades for both English and other less-resourced languages such as Hungarian. However, a large majority of them solely focus on achieving high scores on artificial benchmarks and ignore the importance of practical usability.

In this paper we introduce \huspacy, an industry-strength Hungarian text processing pipeline capable of parsing and tagging texts with high accuracy on limited computational resources. Our system is built upon \spacy’s\footnote{\url{https://spacy.io/}} NLP components, which means that it is fast, has a rich ecosystem of NLP applications and extensions, comes with extensive documentation and a well-known API. 

First, we give an overview of the underlying models, then rigorous evaluation is presented using various datasets. Finally, experiments are presented confirming that \huspacy{} has high accuracy in many subtasks while maintaining resource-efficient prediction capabilities.

\section{Background}

\subsection{Demands for a language processing pipeline in the 2020s}
Starting from the release of the Penn Treebank \citep{penntreebank} in 1992, the research community developed language processing tools for particular tasks, like tokenization, part-of-speech tagging etc. These tools are usually run in a sequence and form a pipeline. In the 2000s, many language-specific corpora and treebanks were developed along with such pipelines. Hungarian was among the best supported languages \citep{metanet} ten years ago.

In the early 2010s, Universal PoS \citep{udpos} and Universal Dependency \citep{univedepv1} labeling schemata were developed with the goals of ''cross-linguistically consistent treebank annotation for many languages'' and ''facilitating multilingual parser development, cross-lingual learning, and parsing research from a language typology perspective.'' Many language-specific pipelines changed their representations to these universal annotation schema, but most of them stayed in their own software architecture. Industrial NLP applications are frequently multi-lingual, i.e. the same NLP task has to be solved in several languages. The demand for standardization over languages is high in commercial partners. Beyond universal PoS and dependency annotations, companies who are not NLP experts but want to apply language processing tools prefer multilingual software frameworks and make business decisions to support Hungarian, based on the availability in multilingual frameworks.

The last five years of NLP are dominated by neural language models (NLM) and the applications based on them \citep{nlp-trends}. Academic research has introduced various deep learning methods outperforming previous state of the art in many areas. Such systems usually employ a single neural network providing end-to-end NLP solutions without the need for specific pipeline steps. Also, several pre-trained multilingual NLMs are becoming available which provide standardized solutions for many languages. Regardless of the multilinguality and the high accuracy of deep learning solutions, a lot of critiques have been raised by real-world industrial NLP projects recently.

These are as follows: deep learning solutions often require far more computational resources compared to classic solutions. They heavily rely on GPU acceleration along with significant memory consumption. What is more, their running costs are usually 10 or even 100 times higher than that of alternative solutions. These drawbacks are questioning the commercial return of the accuracy gain. We also note that modern NLP pipelines consist of static word embedding representations and use deep learning for individual pipeline steps as well, hence, the advantage of large neural end-to-end systems might be very small.

Another industrial demand about language processing systems is to provide human-readable output. Most of the industrial applications are fully or partially rule-based solutions, as (enough) training data for a pure machine learning solution is not available. And there is no free lunch! Each and every real-world application has its own requirements. Rule-based components of these real-world applications require language-specific representations which can be used for defining rules. Such human-readable representations consist of tokens, lemmata, part-of-speech tags, morphological features, dependency parse trees and named entities. Static word embeddings are often integral parts of industrial applications, as many practical algorithms (e.g. semantic textual similarity methods) heavily rely on them.

\subsection{Requirements for industrial-strength language processing pipelines}

Considering decades of experience of practical NLP applications, developing an “industrial-strength” text processing system is a challenging task. First of all, such a tool should tackle the most important text preprocessing tasks including tokenization, sentence splitting, PoS tagging, lemmatization, dependency parsing, named entity recognition and word embedding representation. 

Next, the application has to be accurate enough for real world scenarios while it should be resource conscious at the same time. Furthermore, an industry focused system should be developer friendly, customizable and easy-to-integrate, as NLP modules are integral parts of a larger system in practical applications. These requirements imply that solid documentation should be available as well. Moreover, it is often desired that the underlying machine learning model(s) should be reproducible and controllable.

Last but not least, modern NLP applications are usually multilingual, thus compatibility with international annotation standards  \citep{udpos,univedepv1,UniDep,unimorph} is necessary. Moreover, it is also preferred to be easily usable through a well-known multilingual toolkit.

\subsection{The landscape of Hungarian language processing pipelines}

Up until recently, only a few text processing applications were focused on meeting these criteria even for English. When spaCy was released in 2015 \citep{spacy}, it was one one of the first tools targeting industrial applications mainly. Its authors created an unprecedented tool which offers near state-of-the-art accuracy while being an order of magnitude faster than other tools available. SpaCy also comes with an intuitive API, has detailed documentation, and also fits well into the Python ecosystem of machine learning tools. What is more, it is easily deployable and offers built-in syntax and entity visualization tools as well.

The landscape of the Hungarian text processing systems is similar to that of English before the ''industrial NLP revolution''. There are a number of standalone text analysis tools\footnote{cf. \url{https://github.com/oroszgy/awesome-hungarian-nlp}} \citep{metanet} capable of performing individual text processing tasks, but they often do not play well with each other. 

In contrast, there are only a few attempts at providing industrial Hungarian pipelines. One of them is \magyarlanc{} \citep{magyaralanc} which is a Java based system consisting of state-of-the-art pipeline steps, which were adapted and extended from various libraries. It was designed to serve industrial applications, what is more, a lot of effort has been made on software quality, speed, memory efficiency and customizability. It performs tokenization, sentence boundary detection (SBD), PoS tagging, lemmatization and dependency parsing, but lacks entity recognition and word embeddings. It uses the version 1 of the Universal Dependency (UD) annotations. Although the tool is still used in real world commercial applications, it is not maintained for years.

There is only one other attempt to provide a unified framework for Hungarian text processing tools: \emtsv{} \citep{emtsv1, emtsv2, emtsv3} (and its predecessor e-magyar \citep{emagyar1, emagyar2}) is a result of a multi-institute collaboration project aiming to integrate existing NLP toolkits into a single application. Unfortunately, neither computational efficiency nor developer ergonomics were amongst the main goals of the project. Although \emtsv{} can yield Universal morphosyntactic annotations through conversion, it is rather inaccurate. What is more, it is not designed to efficiently deal with word vectors, therefore no such facility is available in the system.

Talking of Hungarian-specific pipelines, we must mention the contenders of the recent multilingual CoNLL text parsing competitions \citep{conll-2017, conll-2018}. There were numerous  submissions, but \stanza{} \citep{stanza} and \udpipe{} \citep{udpipe} are by far the most popular freely available off-the-shelf applications. These tools provide morphological and syntactical analysis of raw texts for many languages, but lack entity annotations. Accuracy scores vary across tools, but all of them are limited by the small size of the publicly available UD annotated gold standard corpora.

\newlength{\lsys}
\settowidth{\lsys}{embeddings}
\begin{table}
\begin{center}
\begin{tabular}{
    l<{\hspace{1em}}
  >{\centering\arraybackslash}m{\lsys}
  >{\centering\arraybackslash}m{\lsys}
  >{\centering\arraybackslash}m{\lsys}
  >{\centering\arraybackslash}m{\lsys}
  >{\centering\arraybackslash}m{\lsys}
}
\toprule
           & NER & 
           \begin{tabular}{@{}c@{}}Word  \\ embeddings\end{tabular} &
           \begin{tabular}{@{}c@{}}High \\ throughput\end{tabular} & 
           \begin{tabular}{@{}c@{}}Part of a \\ multilingual \\ pipeline\end{tabular}
             &
             \begin{tabular}{@{}c@{}}Free for \\ commercial \\ usage\end{tabular} \\
\midrule
\magyarlanc{} & -                       & -                                   & \checkmark                                   & -                                                   & \checkmark                                             \\
\emtsv{}      & \checkmark                       & -                                   & -                                   & -                                                   & -                                             \\
\udpipe{}     & -                       & -                                   & \checkmark                                   & \checkmark                                                   & -                                             \\
\stanza{}     & \checkmark                       & \checkmark                                   & -                                   & \checkmark                                                   & \checkmark                                            
\\
\bottomrule
\end{tabular}
\vspace{1em}
\caption{Hungarian language processing pipelines evaluated with regards of requirements of industrial applicability}
\label{table:systems}
\end{center}
\vspace{-2em}
\end{table}

Table \ref{table:systems} summarizes the landscape of the most important Hungarian language processing pipelines and show how they meet the requirements of today’s NLP applications. We must note that \emtsv{} does not have any restriction for using it in commercial applications, although some of its most important components have very restrictive licenses (e.g. \texttt{emMorph}). All in all, we can say that none of them is easily applicable in industrial settings.

We present \huspacy{}, a new industry-ready Hungarian natural language processing toolkit.
It provides all the aforementioned basic text processing modules with high accuracy. The underlying models are optimized to be light on memory consumption and CPU usage. The presented tool is open source\footnote{\url{https://github.com/huspacy/huspacy}} and is available under the permissive CC-BY-SA-4.0 license. Our system is built on top of spaCy’s infrastructure, thus extensive documentation, debugging tools, an ergonomic API and a flourishing ecosystem are already provided.

\section{HuSpaCy internals}

This section introduces the NLP algorithms behind the presented tool. As our system is built on spaCy’s architecture, we mainly relied on its symbolic and ML-based text processing infrastructure. The following paragraphs give a high-level overview of the framework utilized and also describes the contributions of this work.

\subsection{Tokenization}

\huspacy{} builds on spaCy\textquotesingle s \citep{spacy-tokenization} tokenization infrastructure which works as follows: first the input text is split on whitespaces, then token boundaries are identified by splitting prefixing or suffixing character sequences. To make this algorithm viable for Hungarian, we extended it with language specific prefix and suffix splitting rules. Furthermore, we had to deal with the ambiguity of tokens around full stops, thus an extensive abbreviation list has been incorporated to increase the module\textquotesingle s accuracy. During this process we mostly relied on the test cases of \huntoken{} \citep{huntoken} to fine-tune the algorithm.

\subsection{Morphosyntactic tagging, sentence splitting and parsing}
\label{sec:tagging}

Sentence boundaries, dependency parse trees, PoS tags, and the corresponding morphosyntactic features are predicted by a multitask deep learning model of the underlying NLP framework.  SpaCy\textquotesingle s machine learning approach can be summarized as ''embed, encode, attend, predict''  \citep{spacy-neural-model} which our system adapts for its tagging and parsing components.

Tokens are embedded using the concatenation of static (pretrained) word vectors and ones learned during the task-specific training process. 
We use a publicly\footnote{\url{https://github.com/oroszgy/hunlp-resources/releases/tag/webcorpuswiki_word2vec_v0.1}} available 300d word embedding which has been trained on the Hungarian Webcorpus \citep{HunWebCorpus} and a snapshot of the Hungarian Wikipedia with CBOW methodology \citep{word2vec}. 
Task specific word vectors are 256 wide consisting 64 dimensional embeddings of the tokens\textquotesingle{} prefixes, suffixes, shapes and the lowercase forms. To make such computation efficient, feature hashing is extensively applied to all kinds of input strings. 

During the encoding part, vectors are passed through a four deep stacked CNN encoder \citep{cnn} which uses residual connections and is accompanied with maxout pooling\footnote{The pooling step is considered to be the “attention” mechanism.} \citep{spacy-parser2}. 
Efficient prediction is guaranteed by the underlying greedy tagger consisting only of a linear and a softmax layer. As for the dependency parsing, an arc-eager transition system \citep{spacy-parser} is utilized, which shares weights with the tagger model through multitask learning. 

Finally, sentence boundary recognition is formalized as a sequence tagging problem where tokens are tagged with a binary label indicating the first token of a sentence. This component is an integral part of the multitask architecture, thus it also shares its neural model with the parser and the morphosyntactic tagger.

\subsection{Lemmatization}

SpaCy\textquotesingle s default lemmatization model is mainly designed for English. It is not suitable for morphologically complex languages such as Hungarian as it only uses lookup tables. Hence, we decided to look for a more sophisticated solution and adapted the \lemmy{} toolkit \citep{lemmy} which is an open-source Python implementation of the CST rule-learning engine \citep{cst-lemmatizer}. To improve its accuracy we incorporated three minor modifications. First, prefixing numbers of numeric tokens are masked to help the engine in case of inflected numbers. (For example the masked token of `2021-ben’ becomes `0000-ben’.) Second, we enforce lowercasing of sentence starting tokens if they are not proper nouns. Finally, if there are multiple lemma candidates available for a given (word, tag) pair, we pick the one with the highest frequency on the training dataset.

\subsection{Named entity recognition}

SpaCy’s entity recognizer is built on the transition-based parser architecture described in Section \ref{sec:tagging} (similarly to \citet{spacy-ner}). 
However, there are two key differences compared to the system of \citet{spacy-ner}. 
The first is that the set of possible transition actions reflects the BILOU tagging scheme. This trick allows the model to have better discrimination ability between different entity classes, furthermore it makes the learning problem easier. 
Second, the state vector computation includes clues not just from the surrounding words but the tokens of previous entities as well. 
The sequence tagger model uses BILOU tags for encoding entity boundaries and the decoder is built on a greedy softmax layer similar to that of the morphosyntactic tagger.

\section{Experiments and results}
\subsection{Text parsing}
\label{parser-eval}

In order to benchmark \huspacy{}, we performed a series of experiments comparing its performance with the most popular off-the-shelf pipelines available. Evaluation is carried out on the test set of the Hungarian Universal Dependencies Corpus \citep{UniDep} by using the evaluation script of the CoNLL 2018 Shared Task\footnote{\url{https://universaldependencies.org/conll18/conll18_ud_eval.py}}. 



Three popular text processing tools have been selected for comparison. \emtsv{} is a Hungarian specific pipeline integrating state-of-the-art NLP components, \udpipe{} is used as a baseline system in CoNLL competitions, while \stanza{} has high scores on parsing UD corpora. All systems are used  as black boxes meaning they have not been retrained or fine-tuned. 


Up until now, there has only been a single Hungarian corpus \citep{szegedcorpus} having both morphosyntactic and dependency parse annotations. What is more, UD annotations are available only in a rather small subcorpus of it \citep{UniDep}. As PoS and morphosyntactic labels can be transcribed automatically from the Hungarian-specific formalism to UD with high accuracy, additional silver standard data can be utilized to train taggers.

In case of \huspacy{}, we applied a two-step learning strategy
\footnote{Hyperparameters of the models are available in the tool\textquotesingle s repository (tag \texttt{v0.4.2}) as configuration files.} to best utilize all available training data. 
In the first step, the tagger and the SBD components are pre-trained on the whole transcribed SZC\footnote{When we refer to the Szeged Corpus as a training set, we mean all the sentences that are not part of the development or the test set of the Universal Dependencies corpus.}. This is followed by a fine-tuning step on the gold standard UD dataset where dependency annotations are also learned. 
To allow fair comparison with \stanza{} and \udpipe{}, a single step model relying solely on the UD data is also involved in the evaluation. 

For similar reasons, the lemmatizer has been trained with two configurations. First we used only the training set of the Hungarian UD corpus, then we allowed the tool to learn from the whole transcribed Szeged Corpus (except the sentences overlapping with either our test or development sets).

The authors of \stanza{}\footnote{\url{https://stanfordnlp.github.io/stanza/performance.html}} and \udpipe{}\footnote{\url{https://ufal.mff.cuni.cz/udpipe/1/models}} have already published their tools\textquotesingle \space accuracy on the Hungarian UD corpus, however the same is not true for \emtsv{.}
To evaluate the latter toolkit we used the following (default) configuration to produce an UD-compatible output: \texttt{emToken, emMorph, emLem, emTag, emmorph2ud, emDep, emConll}. 
While \emtsv{} can provide parse trees, their annotation schema is not compatible with that of the Universal Dependencies, hence, its output is not evaluable. 
We must also note that comparison with the \emtsv{}\textquotesingle s tagger and lemmatizer might not be fair, as this tool was trained on a different train-test split which might conflict with ours. (There is a high chance that its training data overlaps with the sentences of our test set.)

\newlength{\lsz}
\settowidth{\lsz}{Sentence splitting}
\begin{table}
\begin{center}
\begin{tabular}{
  l<{\hspace{1em}}
  >{\centering\arraybackslash}m{\lsz}
  >{\centering\arraybackslash}m{\lsz}
}
\toprule
              & Tokenization             & Sentence splitting \\
\midrule
\stanza{}        & 99.87\%                  & 97.00\%            \\
\udpipe{}        & 99.80\%                  & 95.90\%            \\
\emtsv{}         & 99.77\%                  & 98.67\%            \\
\huspacy{} (UD)  & \multirow{2}{*}{99.89\%}     & 97.66\%            \\
\huspacy{} (SZC) &                          & 97.54\%            \\  
\bottomrule
\end{tabular}
\vspace{1em}
\caption{Tokenization and sentence boundary detection F1 scores on the test of the Hungarian UD Corpus}
\label{table:tokens}
\end{center}
\vspace{-3em}
\end{table}

F1 scores in Table \ref{table:tokens}  suggest that tokenization is easily handled by all of the systems, although \huspacy{} is marginally better compared to the rest of the tools. 
Sentence boundary detection is a more complex task, where language specific knowledge is necessary. This can be either built into the system (as it is the case with \emtsv{}) or learned by a ML model. Numbers show that both approaches can yield satisfactory SBD components, although the rule-based solution of \emtsv{} stands out followed by the tagging approach of our pipeline.

\newlength{\lm}
\settowidth{\lm}{Morph. acc.}
\begin{table}
\begin{center}
\begin{tabular}{
    l<{\hspace{1em}}
  >{\centering\arraybackslash}m{\lm}
  >{\centering\arraybackslash}m{\lm}
  >{\centering\arraybackslash}m{\lm}
  >{\centering\arraybackslash}m{\lm}
}
\toprule
              & PoS acc. & Morph. acc. & UAS   & LAS   \\
\midrule
\stanza{}        & 96.03\%      & 93.76\%               & 83.62\%   & 78.86\%   \\
\udpipe{} v1     & 90.60\%      & 88.50\%               & 72.80\%   & 67.20\%   \\
\emtsv{}         & 89.19\%      & 89.12\%               & \multicolumn{2}{c}{--} \\
\huspacy{} (UD)  & 94.70\%      & 89.03\%               & 79.03\%   & 73.17\%   \\
\huspacy{} (SZC) & 96.58\%      & 93.23\%               & 79.39\%   & 74.22\%   \\
\bottomrule
\end{tabular}
\vspace{1em}
\caption{Comparison of  tagging accuracy and attachment scores of the benchmarked pipelines on the test set of the Hungarian UD Corpus.}
\label{table:tagging}
\end{center}
\vspace{-3em}
\end{table}

Tagging accuracy and attachment scores are presented in Table \ref{table:tagging}. 
Results show that \stanza{} is a clear winner in dependency parsing while the PoS tagging score of \huspacy{} (the one using additional training data) is the highest one. 
It can be seen that the usage of the extra silver standard data yields better performance for our models both during tagging and dependency parsing. 
\udpipe{} and \emtsv{} have relatively low scores: 
the results of \udpipe{} are not surprising (cf. \cite{conll-2018}), but  \emtsv{}\textquotesingle s scores are unexpected given that it is built upon state-of-the-art morphosyntactic tagging facilities \citep{purepos}. 

\newlength{\llem}
\settowidth{\llem}{Accuracy}
\begin{table}
\begin{center}
\begin{tabular}{
    l<{\hspace{1em}}
  >{\centering\arraybackslash}m{\llem}
}
\toprule
              & Accuracy \\
\midrule
\stanza{}        & 94.25\%    \\
\udpipe{} v1     & 88.50\%    \\
\emtsv{}         & 94.94\%    \\
\huspacy{} (UD)  & 94.82\%    \\
\huspacy{} (SZC) & 95.53\%    \\
\bottomrule
\end{tabular}
\vspace{1em}
\caption{Lemmatization accuracy of NLP pipelines measured on the test set of the Hungarian UD Corpus. \huspacy{} (UD) uses the same setting as its contenders, while \huspacy{} (SZC) builds on the whole Szeged Corpus for training.}
\label{table:lemma}
\end{center}
\vspace{-3em}
\end{table}

Lemmatization results in Table \ref{table:lemma} show that all the systems except \udpipe{} are accurate enough. \huspacy{} trained on the full Szeged Corpus stands out, its score is more than 0.5\% higher than the second best system (\emtsv{}). The best configuration of \huspacy{} scores more than 0.5\% higher than the one trained solely on the UD dataset. 

\subsection{Named entity recognition}

Comparing NER components is not as straightforward as it is for the parsing subtasks. There are multiple evaluation datasets, but there is no consensus between researchers on their usage. NYTK-NerKor (NerKor) \citep{NerKor} is a relatively new corpus consisting of  1 million tokens, while SzegedNER \citep{szeged-ner-corpus} is a 200,000 token subset of the Szeged Corpus.  \citet{nerkor-eval} uses the former dataset to benchmark some of the most popular tools, while previous work mainly rely on the latter one.
 \udpipe{} does not have a NER component, thus we cannot include it in this investigation. As for \emtsv, its entity recognizer was trained using the whole SzegedNER corpus, its comparison against other tools would not be fair.

\huspacy\textquotesingle s entity recognition capabilities are benchmarked in this work on both corpora using the same train-test splits as \citet{szarvas-ner} and \citet{nerkor-eval} suggest. 
As Hungarian entity recognition datasets share the same tagset and rely on similar annotation guides, it is possible to train models using both corpora. In this regards, we follow the work of \citet{nerkor-eval} and evaluate \huspacy{} on the combined corpus as well.
We also include results of previous entity recognition attempts so as to put our results in context. One of the first systems was developed by \cite{szarvas-ner}, which utilizes decision trees for tackling the problem. Next, there is HunTag \citep{huntag, simon-ner}, which is a statistical tagger utilizing a linear model combined with Hidden Markov models. \citet{simon-ner} also showed that it is possible to improve on the F1 score of the base system by incorporating silver standard data. Most recently, \cite{embert} developed an entity recognizer on top of Hungarian BERT models \citep{hubert} achieving state-of-the-art results.

\newlength{\lnersz}
\settowidth{\lnersz}{SzegedNer}

\begin{table}
\begin{center}
\begin{tabular}[t]{
    l<{\hspace{1em}}
  >{\centering\arraybackslash}m{\lnersz}
  >{\centering\arraybackslash}m{\lnersz}
  >{\centering\arraybackslash}m{\lnersz}
}
\toprule
                           & SzegedNer & NerKor & Combined\\
\midrule
\cite{simon-ner}           & 95.06\%   & --      & -- \\
\cite{szarvas-ner}         & 94.77\%\   & --      & -- \\
\embert{}                  & 97.40\%   & 92.09\% & 92.99\% \\
\stanza{}                  & 91.78\%   & 80.53\% & 83.75\% \\
\huspacy{}                 & 95.31\%   & 80.75\% & 83.46\% \\
\bottomrule
\end{tabular}
\vspace{1em}
\caption{Comparison of entity recognition F1 scores on the SzegedNER test set \citep{szarvas-ner}, on the NerKor test set and on the combined test}
\label{table:ner_cmp}
\end{center}
\hspace*{.1\linewidth}
\vspace{-2em}
\end{table}

Table \ref{table:ner_cmp} contains F1 scores of all the entity recognizers mentioned above. 
It can be seen that the BERT-based model achieves the highest scores on all of the datasets with a large margin. However, these models are also well-known for their enormous computational costs. \huspacy{} is the second best contender on SzegedNER, although its performance is on pair with \stanza{} on other datasets. \embert{}\textquotesingle s results are outstanding when NerKor is involved in the comparison. As \citeauthor{nerkor-eval} concludes these measurements are in accordance with similar English NER benchmarks (cf. \cite{stanza}). Pretrained transformer-based models often yield significantly higher performance scores compared to other sequence tagging approaches due to the underlying attention mechanism and the their model\textquotesingle s increased capacity. But there is no free lunch, higher accuracy comes with significantly increased prediction costs. 

The final model of \huspacy{} builds on the weights of a pretrained neural tagger (using the strategy described in Section \ref{parser-eval}) yielding 84.56\% F1 on the combined dataset. This result is a significant improvement compared to \stanza \textquotesingle s score and also confirms the usefulness of additional silver standard training data usage for spaCy\textquotesingle s multitask neural model.

\subsection{Resource usage}
Resource usage such as memory consumption and processing speed is an important aspect of practical text processing systems, thus we benchmarked\footnote{All experiments were performed on a computer having an Intel Core i7-8750H CPU and 16 GB RAM running Ubuntu Linux 20.04 LTS.} text parsing pipelines (cf. Section \ref{parser-eval}) in this respect. In order to have a fair comparison, we configured all systems to perform only tokenization, sentence splitting, PoS tagging, lemmatization and dependency parsing. As timing measurements should ignore model loading times, Stanza and \udpipe{} were used by their Python interfaces, while \emtsv{} was utilized through its REST API. We used the UD test set to measure throughput and peak memory consumption.

\newlength{\lper}
\settowidth{\lper}{Throughput (tokens/sec)}
\begin{table}
\begin{center}
\begin{tabular}{
    l<{\hspace{1em}}
  r>{\centering\arraybackslash}m{\lper}
  r>{\centering\arraybackslash}m{\lper}
}
\toprule
        & Throughput (tokens/sec) & Memory usage (GB) \\
\midrule
\stanza{}  & \num{222}  & \num{0.9} \\
\udpipe{}  & \num{1741} & \num{0.4} \\
\emtsv{}   & \num{122}  & \num{3.9} \\
\huspacy{} & \num{2612} & \num{2.1} \\
\bottomrule
\end{tabular}
\vspace{1em}
\caption{Throughput (measured in tokens/second) and peak memory consumption of benchmarked NLP pipelines.}
\label{table:performance}
\end{center}
\vspace{-2em}
\end{table}

Table \ref{table:performance} 
presents computational efficiency measures suggesting that our system has the highest throughput amongst all the tools. \huspacy{} is almost 50\% faster than \udpipe{}, while producing significantly better parses. As regards \stanza{}, there is a huge tradeoff on having the best dependency parser: it is almost 8 times slower than \udpipe{} and more than 10 times slower compared to our system. 

Memory consumption of the pipelines are acceptable as all of them could easily fit in a modern computer’s RAM. Our tool has the highest memory usage which is due to its 300-dimensional word vectors. In comparison, \stanza{} is the only other tool having word embeddings, but its vectors\textquotesingle{} sizes are limited to 100d.

\section{Conclusions}

We presented \huspacy{}, a new industry-ready Hungarian language processing pipeline that is open source and is freely available. While previous approaches have failed to provide a tool which can be easily used to solve practical text processing problems, our system builds on the solid foundations of an industrial NLP framework. We presented how our toolkit utilizes spaCy\textquotesingle s underlying ML models to provide all the basic language analysis components. We performed various experiments proving that our system has high accuracy in many text processing tasks while using only moderate computation resources.

As results show, the accuracy of \huspacy\textquotesingle s dependency parser needs further improvements. 
Further advancement opportunities lie in fine-tuning the NER model and in using a new neural lemmatizer. 

In summary, this study described a new freely available tool which is suitable for real-world industrial applications.

\section*{Acknowledgements}
The authors would like to thank Dávid Nemeskey and Dániel Lévai for their help in benchmarking \embert{} and \stanza{}.
HuSpaCy research and development is funded by the Ministry of Innovation and Technology NRDI Office within the framework of the Artificial Intelligence National Laboratory Program. 

%
\renewcommand\bibname{\refname}
\renewcommand\bibname{References}
\bibliographystyle{splncsnat_en}
\bibliography{main}

\end{document}